\documentclass{SCIS2024}
\usepackage{type1cm}
\usepackage{array}
\usepackage{textcomp}
\usepackage{stfloats}
\usepackage{multirow}
\usepackage{xspace}
\newcolumntype{L}[1]{>{\raggedright\arraybackslash}m{#1}}
\newcolumntype{C}[1]{>{\centering\arraybackslash}m{#1}}
\newcolumntype{R}[1]{>{\raggedleft\arraybackslash}m{#1}}

\usepackage[capitalize]{cleveref}
\crefname{section}{Sec.}{Secs.}
\Crefname{section}{Section}{Sections}
\Crefname{table}{Table}{Tables}
\crefname{table}{Tab.}{Tabs.}
\makeatletter
\DeclareRobustCommand\onedot{\futurelet\@let@token\@onedot}
\def\@onedot{\ifx\@let@token.\else.\null\fi\xspace}

\def\etc{\emph{etc}\onedot} 
 
\def\etal{\emph{et al}\onedot}
\makeatother

\begin{document}
\ArticleType{RESEARCH PAPER}
\Year{2022}
\Month{}
\Vol{}
\No{}
\DOI{}
\ArtNo{}
\ReceiveDate{}
\ReviseDate{}
\AcceptDate{}
\OnlineDate{}

\title{Towards Imbalanced Motion: Part-Decoupling Network for Video Portrait Segmentation}{Towards Imbalanced Motion: Part-Decoupling Network for Video Portrait Segmentation}

\author[1]{Tianshu YU}{}
\author[2]{Changqun XIA}{{xiachq@pcl.ac.cn}}
\author[1,2]{Jia LI}{{jiali@buaa.edu.cn}}

\AuthorMark{Tianshu Yu, Changqun Xia, Jia Li}

\AuthorCitation{Tianshu Yu, Changqun Xia, Jia Li}

\address[1]{State Key Laboratory of Virtual Reality Technology and Systems, School of Computer Science and Engineering,\\ Beihang University, Beijing {\rm 100191}, China}
\address[2]{Peng Cheng Laboratory, Shenzhen {\rm 518055}, China}

\abstract{Video portrait segmentation (VPS), aiming at segmenting prominent foreground portraits from video frames, has received much attention in recent years. However, simplicity of existing VPS datasets leads to a limitation on extensive research of the task. In this work, we propose a new intricate large-scale Multi-scene Video Portrait Segmentation dataset MVPS consisting of 101 video clips in 7 scenario categories, in which 10,843 sampled frames are finely annotated at pixel level. The dataset has diverse scenes and complicated background environments, which is the most complex dataset in VPS to our best knowledge. Through the observation of a large number of videos with portraits during dataset construction, we find that due to the joint structure of human body, motion of portraits is part-associated, which leads that different parts are relatively independent in motion. That is, motion of different parts of the portraits is imbalanced. Towards this imbalance, an intuitive and reasonable idea is that different motion states in portraits can be better exploited by decoupling the portraits into parts. To achieve this, we propose a Part-Decoupling Network (PDNet) for video portrait segmentation. Specifically, an Inter-frame Part-Discriminated Attention (IPDA) module is proposed which unsupervisedly segments portrait into parts and utilizes different attentiveness on discriminative features specified to each different part. In this way, appropriate attention can be imposed to portrait parts with imbalanced motion to extract part-discriminated correlations, so that the portraits can be segmented more accurately. Experimental results demonstrate that our method achieves leading performance with the comparison to state-of-the-art methods.}


\maketitle

\section{Introduction}
\label{sec:intro}
Video portrait segmentation (VPS)~\cite{PVS}, which aims to discover and separate prominent foreground portraits from video frames, has drawn much interest due to a great number of different application scenarios in video creation such as background replacement~\cite{Background2021}, portrait transformation~\cite{EG1800}, \etc.

Currently, several video segmentation datasets with portraits have been proposed, like DAVIS~\cite{DAVIS}, PVSD2.5K~\cite{PVS}, PP-HumanSeg14K~\cite{PP-HumanSeg}, \etc. In DAVIS, video clips with human as foreground objects are all captured in outdoor scenes. PVSD2.5K consists of only 2,530 annotated frames, and there is only one prominent person in each video clip. PP-HumanSeg14K only contains videos in remote conference, in which background environments are quite simple. Although recent unsupervised VOS methods~\cite{COSNet, AGNN, MATNet, GraphMemVOS, F2Net, RTNet, FSNet, AMC-Net, HFAN, FEM-Net, IMCNet} and a specialized VPS method~\cite{PVS} based on existing datasets achieve good performance, due to the uniformity of scenarios and the monotonousness of background environments, the simplicity of these datasets leads to weak robustness of models to handle complex situations in practical applications.

\begin{figure}[t]
    \centering
    \includegraphics[width=\linewidth]{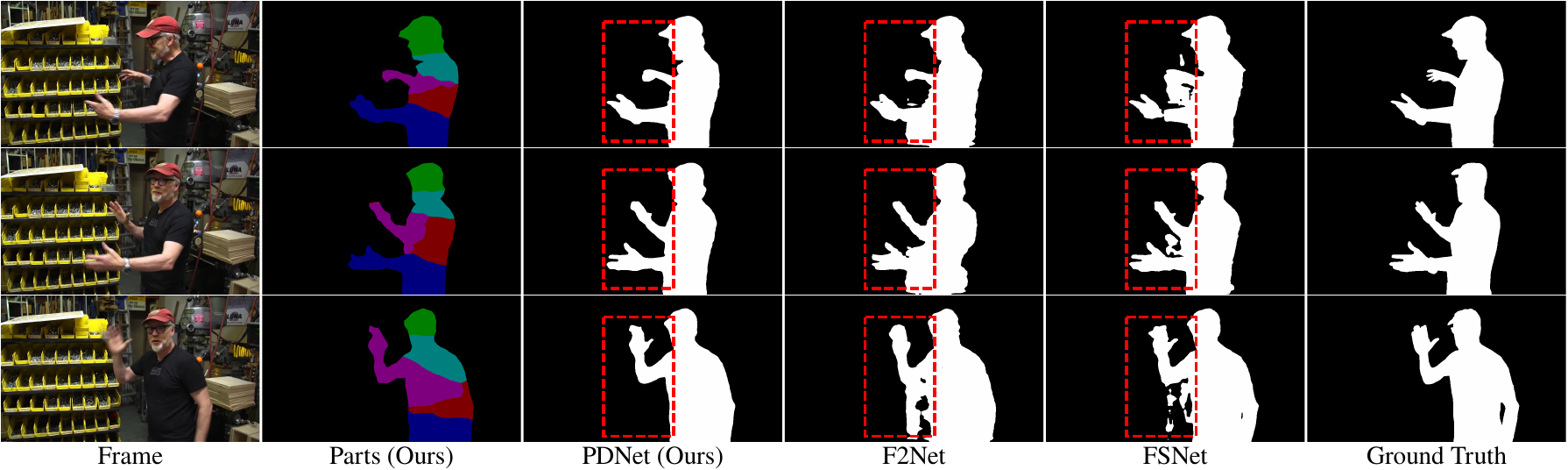}
    \caption{Motion imbalance between arms and main body. Due to the joint structure of human body, different parts in portraits have relatively independent motion state.}
    \label{fig:first}
\end{figure}

In this work, we propose a new intricate large-scale Multi-scene Video Portrait Segmentation dataset MVPS. The dataset consists of 101 video clips in 7 categories of different scenarios including entertainment, indoor handwork, interview, lecture, news, outdoor activity and online shopping which appear frequently on the Internet. Portraits in these scenarios are in different poses and gestures, and background scenes are also diverse. These complications are much closer to practical application scenarios on the Internet. We sample 53,923 frames from these video clips in total, where 10,843 of them are finely annotated. It is currently the most complex dataset for video portrait segmentation to our best knowledge. 

Based on the observation of a large number of videos with portraits during dataset construction, we find that different from other moving objects like vehicles and airplanes which motion state is consistent for the whole object, motion of portraits is imbalanced due to the joint structure of human body. As shown in \cref{fig:first}, motion of the arms in the red box is independent from that of the main body, which leads to inaccurate prediction near the arm with a greater range of motion in existing methods such as F2Net~\cite{F2Net} and FSNet~\cite{FSNet}. Due to this part-associated imbalance, utilizing same attentiveness on motion of different parts may cause imprecise location and segmentation. Since part-discriminated features can be extracted through unsupervised part segmentation~\cite{SCOPS, PartDAS, FRRG, Mix-ViT, SS3DR, Zhao2023}, an intuitive idea is to introduce this discriminated part cue to extract correlations of imbalanced motion.

To this end, we propose a novel Part-Decoupling Network (PDNet) for video portrait segmentation, which captures the motion correlations of different parts respectively. Specifically, we propose an Inter-frame Part-Discriminated Attention (IPDA) module, which decouples imbalanced integral portrait motion into independent part motion. This module unsupervisedly segments portrait parts of both target frame and reference frame, utilizes cross-attention operation between the same part in different frames to obtain part-discriminated motion features, and finally assembles them to generate global motion features based on masks of these parts predicted by the module. Effectiveness of proposed method is demonstrated in our experiments.

Our major contributions can be summarized as follows:

\begin{itemize}
  \item We propose a new intricate large-scale Multi-scene Video Portrait Segmentation dataset MVPS with 7 categories of scenarios, which is the most complex video portrait segmentation dataset currrently.
  \item We explore the imbalance of portrait motion and propose a Part-Decoupling Network, which leverages the part-associated imbalance of portrait motion and separately captures this imbalanced motion from different parts explicitly.
  \item We propose an Inter-frame Part-Discriminated Attention module to extract discriminative part-level correlations of imbalanced motion.
  \item Experimental results demonstrate that our method achieves superior performance compared to state-of-the-art methods.
\end{itemize}

\section{Related work}
In this paper, we mainly discuss a new VPS dataset and a newly designed VPS method based on the characteristics summarized from the observation during dataset construction. In addition, we use part information as a clue and perceive the imbalanced motion of portraits in videos. We provide an introduction to relevant works.

\subsection{Video portrait datasets}
\label{sec:ps}
Existing segmentation works based on images have achieved great success on salient object detection~\cite{PGNet, CTDNet, PFSNet, ICON, HLSOD, DNTDF, RLLNet, WhuCoS}, image portrait segmentation~\cite{EG1800, PortraitNet, SINet, HISE}, \etc. With the increasing demand for video applications, video portrait segmentation has received much attention by researchers in recent years. Chu \etal~\cite{PP-HumanSeg} propose a large-scale video portrait segmentation dataset PP-HumanSeg14K for remote conference scene with 14,117 annotated frames, in which portraits are always in half-body and there is usually no changes in shooting distance and angle. Wang \etal~\cite{PVS} introduce a dataset PVSD2.5K for VPS which includes 2,530 annotated frames, where there is only one prominent person in each video clip. However, existing datasets only involve some simple scenarios, which leads to poor generalization ability when applied in practical application situations. The lack of datasets with more complex scenarios limits extensive research of the task.

\subsection{Video portrait segmentation}
VPS can be regarded as a special circumstance of video object segmentation (VOS) where the object category is fixed as human~\cite{PVS}. As a hot topic, VPS has been studied by many researchers~\cite{HumanCRF, PersonSegVid, JointPersonSegIdVid, SurvVideoSeg} and regarded as an independent task. A recent work~\cite{PVS} study the VPS task under the settings of unsupervised VOS, which can provide a better assessment of the comprehensive performance of different video clips, so we follow these settings in our work.

As the prominent foreground objects are not marked in VPS, researches have made efforts to locate prominent foreground objects in video frames based on appearance and motion cues. Some works~\cite{PDB, RVOS, AGS, DBSNet} apply recurrent structures to extract temporal-correlated spatial features, which may accumulate interference signals caused by complex backgrounds. Other researches~\cite{LSMO, EpO, MATNet, RTNet, FSNet, AMC-Net, MS-APS, HFAN, TransportNet, FEM-Net, PSNet} utilize optical flow to obtain motion-based temporal features. These methods usually extract spatial and temporal features in two different branches and fuse them, which explicitly exploit appearance and motion cues separately. Some other methods~\cite{PVS, AGNN, COSNet, AnDiff, GraphMemVOS, F2Net, WCS-Net, IMP, IMCNet, FTAL, DMVOS} establish correlations between features extracted from the target frame and reference frames, which is conducive to enhancing temporal global understanding. However, distinct from other common objects like cars, motion states of different parts of portraits are usually different. Many existing methods regard the prominent object as a whole and extract motion information uniformly, which results in incorrect handling of parts with different motion states from the main body. How to treat parts with different motion states separately is another focus of our work.

Note that portrait matting is another distinct task outputting alpha mattes usually guided by trimap or background image, which focuses on obtaining fine boundaries, while video portrait segmentation focuses on locating and segmenting portraits with motion continuity~\cite{PVS}.

\subsection{Unsupervised part segmentation}
Unsupervised part segmentation provides a paradigm to represent object semantics robust to environment and object gesture without any part-level mask annotation, which is first proposed in~\cite{SCOPS}. This work utilizes self-supervision losses based on saliency map to obtain pixel-wise co-part segmentation results among images of a specific object category, which is robust to instance and posture differences. In~\cite{PartDAS}, a method is proposed to learn part-level invariant features between origin image and spatial-transformed image through image reconstruction to generate parts that are stable to spatial displacement. Recent works also utilize part semantics generated by unsupervised part segmentation on several different computer vision tasks related to regional property, such as fine-grained recognition~\cite{FRRG, Mix-ViT}, 3D reconstruction~\cite{SS3DR}, \etc. In our work, we utilize unsupervised part segmentation to extract correlations of part-discriminated imbalanced motion for VPS.

Note that the final output of our work is not for the segmentation of human parts but the whole portraits, while part cue is introduced as a guidance to promote the effectiveness of VPS.

\subsection{Video motion perception}
By perceiving imbalanced human motion in videos, the accuracy of portrait localization and segmentation can be improved. Some approaches~\cite{BackTAL, SODA, BaSNet} suppress background frame signals to enhance the perceptual response to action frames. Some other methods~\cite{ECM, TriDet} capture representation responses in the temporal domain to detect potential motion in videos. In our work, we achieve the perception of imbalanced motion in videos by extracting part-discriminated motion correlation from decoupled parts.

\begin{figure}[t]
  \centering
  \subfloat[Number of video clips in each of 7 scenario categories.]{
    \centering
    \includegraphics[height=5cm]{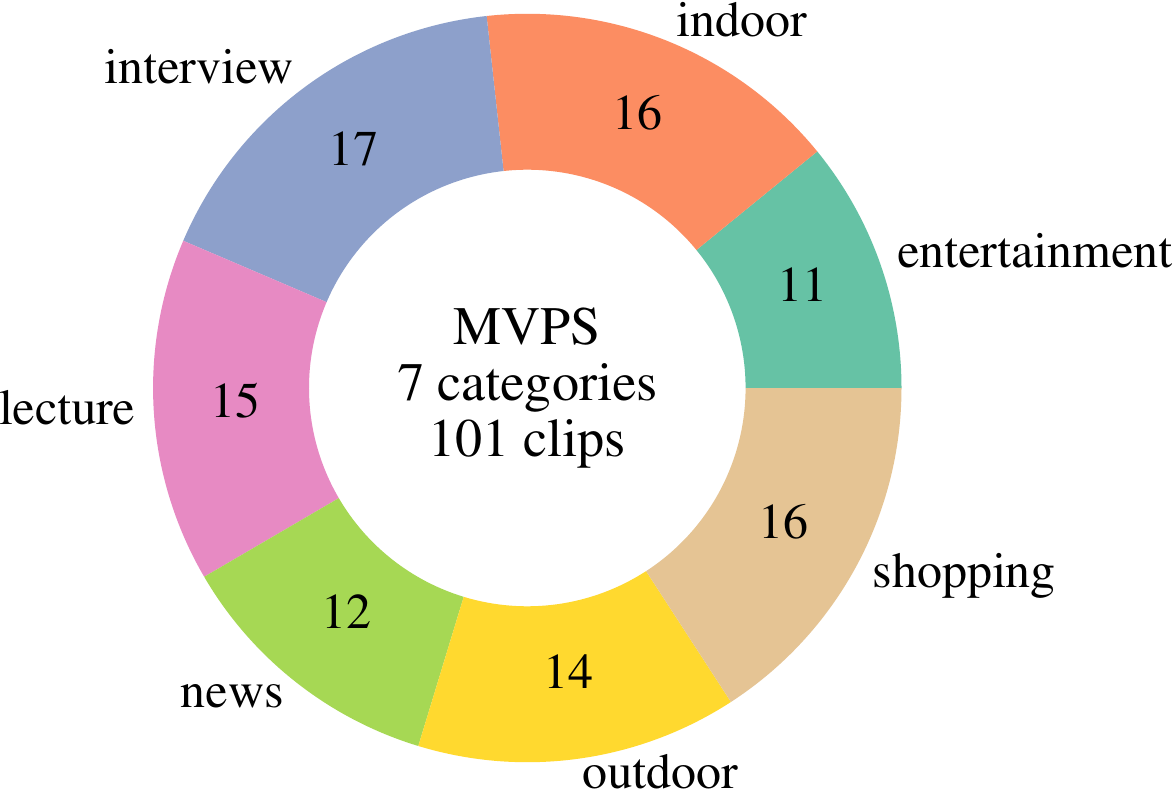}
    \label{fig:data-a}
  }
  \hfill
  \subfloat[Examples for each scenario category.]{
    \centering
    \includegraphics[height=5cm]{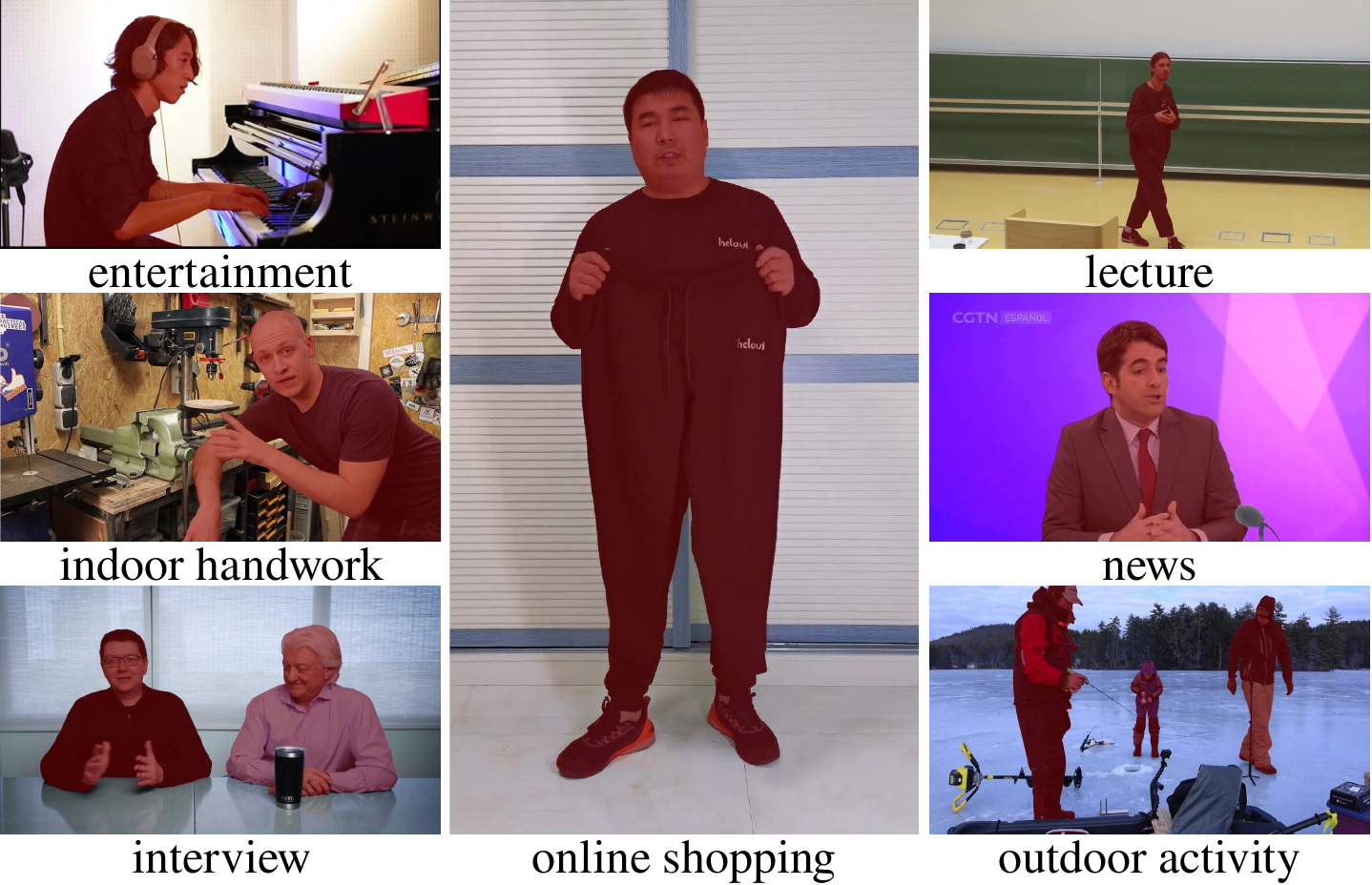}
    \label{fig:data-show}
  }
  \caption{An overview of our MVPS dataset.}
  \label{fig:data}
\end{figure}

\begin{figure}[t]
  \centering
  \subfloat[Distribution of GLCM entropy in MVPS dataset.]{
    \centering
    \includegraphics[height=4.2cm]{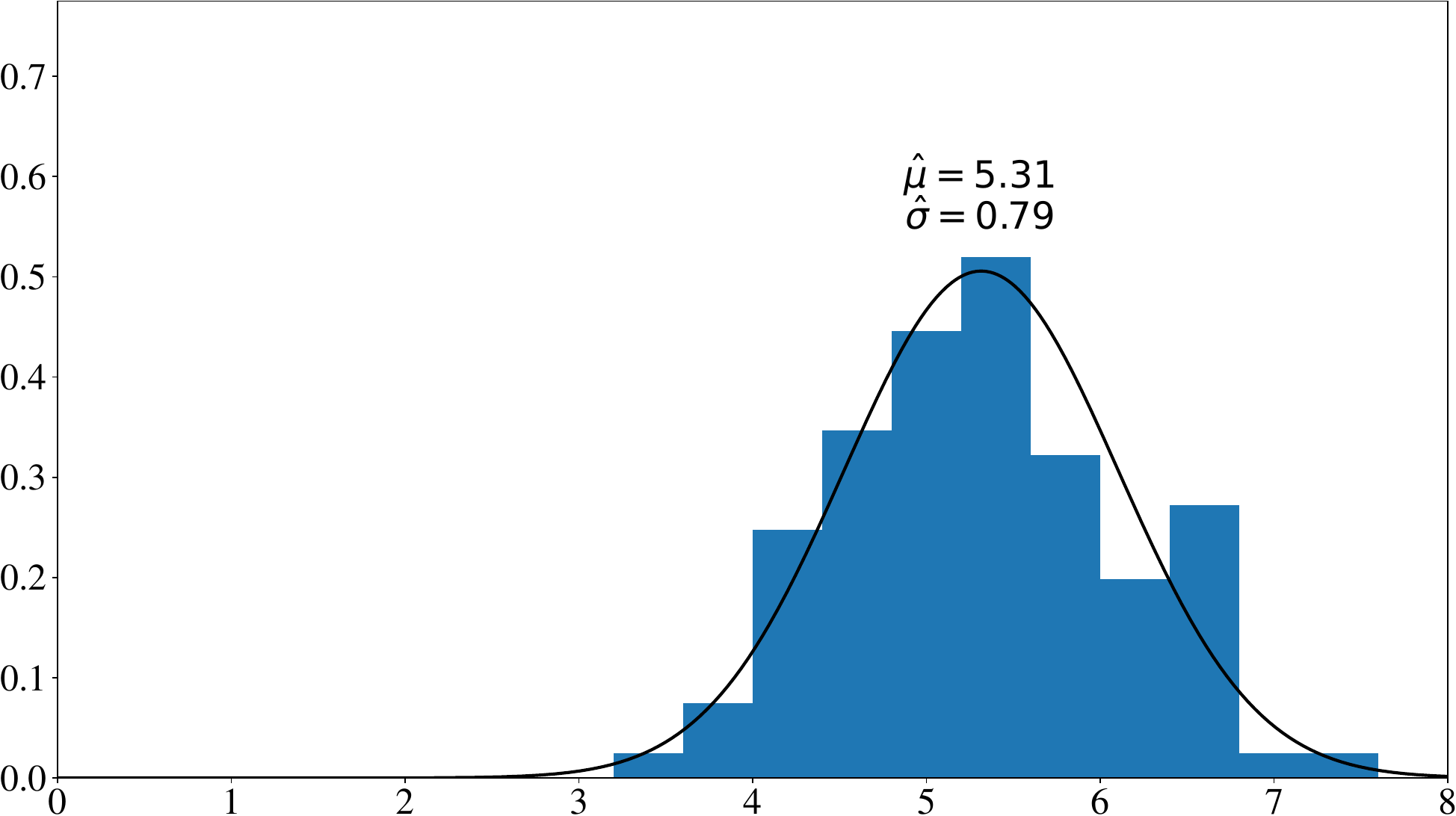}
    \label{fig:glcm-entropy-MVPS}
  }
  \hfill
  \subfloat[Distribution of GLCM entropy in PP-HumanSeg14K dataset.]{
    \centering
    \includegraphics[height=4.2cm]{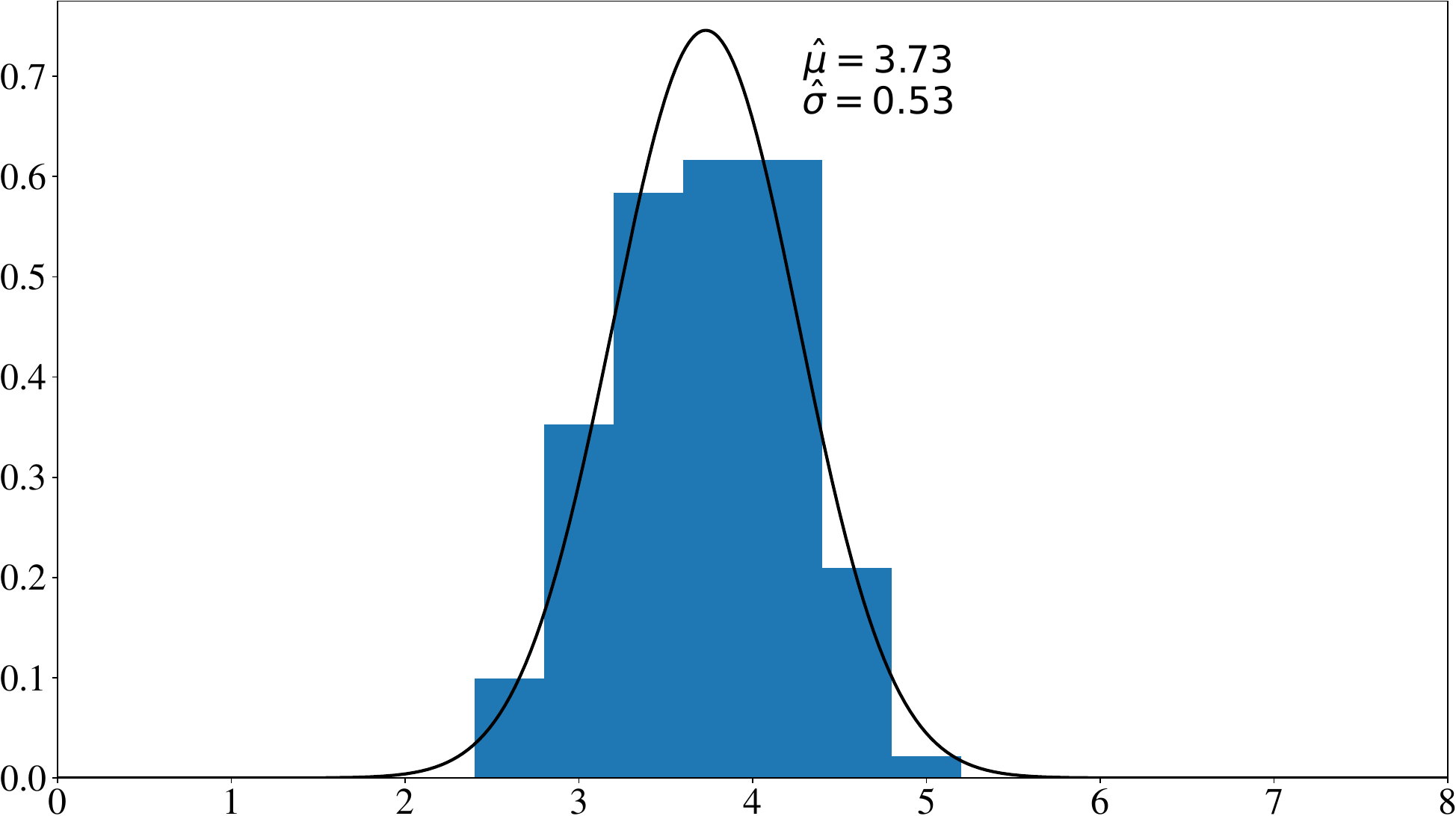}
    \label{fig:glcm-entropy-pp}
  }
  \caption{Frequency distributions of GLCM entropy of video clips in MVPS and PP-HumanSeg14K~\cite{PP-HumanSeg} datasets.}
  \label{fig:glcm-entropy}
\end{figure}

\section{MVPS dataset}
In VPS, the richness of background scenes and the complexity of portrait motion in a dataset have a significant impact on the robustness of segmentation models. Existing video segmentation datasets with portraits~\cite{FBMS, DAVIS, DAVSOD, Youtube-VOS, PVS, PP-HumanSeg} cannot meet the requirements of complexity in scenarios. In order to alleviate the large generalization gap of VPS methods, we construct a new intricate large-scale Multi-scene Video Portrait Segmentation dataset MVPS with 7 categories of scenarios, which is the most complex video portrait segmentation dataset available.

\subsection{Dataset construction}
The diversity of scenes is emphasized during data collection. We first collect initial candidate videos from the public Internet according to the following rules: (1) there is at least one foreground portrait in the video; (2) the number of foreground portraits in videos is different; (3) the background scene of these videos are varied, containing many common indoor and outdoor environments; (4) the videos are clear enough, of which the pixel number on the short side must not be less than 720; (5) videos captured both horizontally and vertically are included. From these candidate videos, we select available video clips which last at least 6 seconds and include at least one foreground portrait with obvious motion and no camera cut. Finally, we obtain 101 video clips for MVPS dataset which are divided into 7 categories of scenarios: entertainment, indoor handwork, interview, lecture, news, outdoor activity and online shopping. Then we sample original frames from these video clips at 30 FPS and obtain 53,923 frames in total, which are finely annotated with portrait masks at pixel-level every 5 frames by 8 experienced annotators. All the annotations are cross-checked to ensure accuracy. Totally there are 10,843 frames annotated in MVPS. For convenience of use, we randomly divide the dataset into training set and test set. The training set contains 61 video clips and 6,732 annotated frames, while the test set contains 40 video clips and 4,111 annotated frames. All frames in MVPS have a high resolution with pixel number of the short side between 720 and 1080. Same as DAVIS~\cite{DAVIS}, we also provide a 480p-resolution version of our MVPS dataset for easier training and evaluation. Note that all videos are available on the public Internet, and our annotations will be publicly available for academic research only.

\subsection{Dataset analysis}
\paragraph{Scenario distribution}
Our MVPS dataset includes videos of 7 scenario categories, where the number of clips in each category distributes quite evenly as shown in \cref{fig:data-a}. Examples for each scenario category are shown in \cref{fig:data-show}. It can be seen that not only the scenario category has a complex distribution, but the background scenes are also significantly different inside each category. In the category ``outdoor activity'', for example, there are videos captured in the jungle, on the river boat, in the snowfield, \etc. Meanwhile, both videos shoted during the day and night are included. Some videos in the dataset contain subtitle occlusion, which is common in practical application scenarios.

\paragraph{Complexity}
To evaluate the complexity of image scenes, we utilize Gray Level Co-occurrence Matrix (GLCM) entropy according to~\cite{DatasetComplexity}. We calculate the average GLCM entropy of frames in each video clip on the typical PP-HumanSeg14K~\cite{PP-HumanSeg} dataset and our MVPS dataset, which frequency distributions are shown in \cref{fig:glcm-entropy}. From the comparison results it can be seen that our MVPS dataset has not only a larger mean but also a larger value range of GLCM entropy, while its values in PP-HumanSeg14K are significantly smaller and more concentrated. This indicates the complexity of scenes in our MVPS dataset.

\begin{figure}[t]
    \centering
    \includegraphics[height=4.5cm]{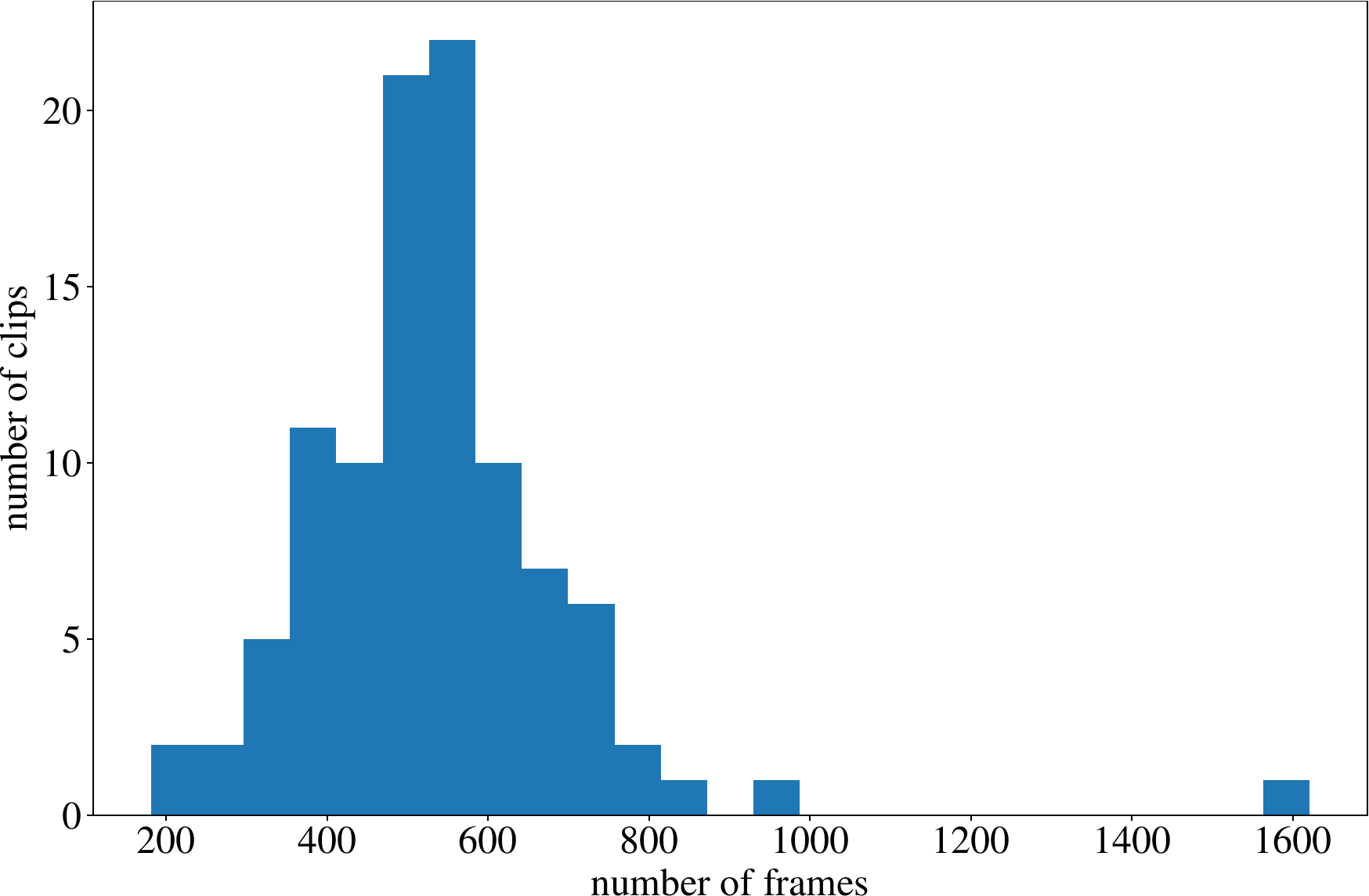}
    \caption{Number of frames in each video clip.}
    \label{fig:data-length}
\end{figure}

\paragraph{Duration}
Duration is also an important factor affecting the complexity of portrait motion. We show the distribution of number of frames in each video clip in \cref{fig:data-length}. Most of the video clips in MVPS have a number of frames between 300 and 900, corresponding to a time length between 10 and 30 seconds. In long video clips, there is more complicated portrait motion. We have not selected clips less than 6 seconds, in which the actions of figures are not abundant enough.

In conclusion, our MVPS dataset contains a variety of intricate scenes and complex portrait motion, which is the most complex dataset in VPS to our best knowledge. It addresses the lack of intricate large-scale multi-scene VPS datasets, which may promote extensive researches on VPS. Our annotations will be publicly available under request for academic research only.

\section{Method}
 Through observation of a large number of videos during dataset construction, motion of portraits is usually imbalanced due to the joint structure of human body, where motion of different parts of portraits is relatively independent. This motion imbalance leads to inaccurate prediction near the parts with a greater range of motion in existing methods, as shown in \cref{fig:first}. Towards this imbalance, we propose a Part-Decoupling Network (PDNet) for video portrait segmentation, which leverages part cue to separately capture part-associated motion in VPS. The network architecture is introduced in~\cref{sec:PDNet}. In our network, an Inter-frame Part-Discriminated Attention (IPDA) module is proposed to extract motion information specified to each part, which is introduced in~\cref{sec:IPDA}.

\begin{figure}[t]
  \centering
  \includegraphics[width=\linewidth]{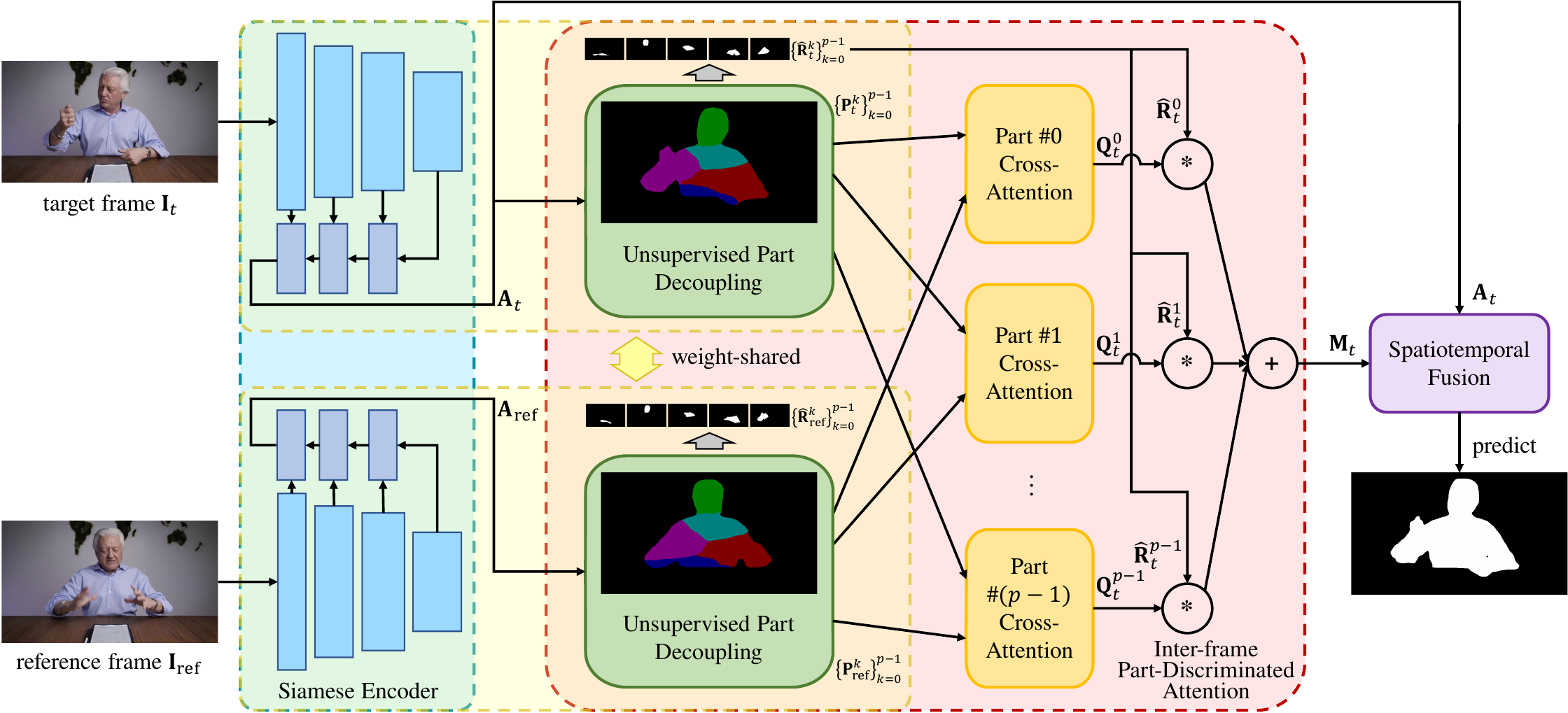}
  \caption{Overview framework of our proposed PDNet.}
  \label{fig:PDNet}
\end{figure}

\subsection{Part-decoupling network}
\label{sec:PDNet}
The architecture of proposed PDNet is shown in~\cref{fig:PDNet}. Similar to several unsupervised VOS methods~\cite{AnDiff, F2Net}, the first frame is used as a reference frame in our network to distill motion information. In the Siamese encoder, we use ResNet-50~\cite{ResNet} as backbone to extract spatial features of both target frame $\mathbf{I}_t$ and reference frame $\mathbf{I}_\text{ref}$, and then take the advantages of FPN~\cite{FPN} structure to fuse spatial semantic features and detailed features from deep to shallow. The whole encoder is weight-shared between the two frames, so that consistent appearance features $\mathbf{A}_t$ and $\mathbf{A}_\text{ref}$ are obtained for the same persons in different frames.

Towards the motion imbalance of portraits, we utilize our proposed Inter-frame Part-Discriminated Attention (IPDA) module to unsupervisedly decouple $\mathbf{A}_t$ and $\mathbf{A}_\text{ref}$ into part-attached features, and utilize cross-attention operation on each part to extract part-discriminated motion correlation. These correlation features are then assembled to generate motion features $\mathbf{M}_t$ of the target frame.

After obtaining both appearance features $\mathbf{A}_t$ and motion features $\mathbf{M}_t$ of the target frame, we utilize a spatiotemporal fusion module to fuse these two features, which is introduced in \cref{sec:fusion}. At last, the final result is predicted from the fused features.

\subsection{Inter-frame part-discriminated attention}
\label{sec:IPDA}
Due to the motion imbalance, our proposed Inter-frame Part-Discriminated Attention (IPDA) module extracts motion correlation from reference frame to target frame separately for each part of portraits, which consists of three stages: unsupervised part decoupling stage, correlation extraction stage and correlation assembly stage.

\begin{figure}[t]
  \centering
  \subfloat[Pipeline of unsupervised part decoupling in IPDA.]{
    \centering
    \includegraphics[height=3.3cm]{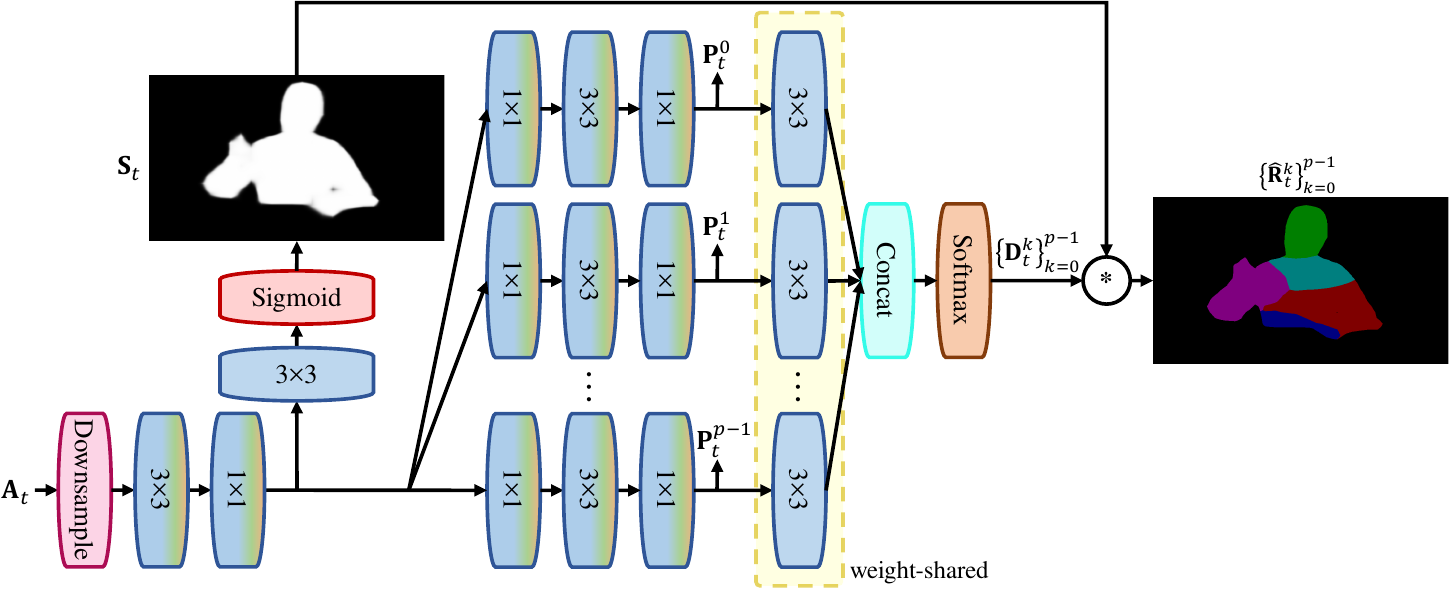}
    \label{fig:PDStage}
  }
  \hfill
  \subfloat[Cross-attention for correlation extraction in IPDA.]{
    \centering
    \includegraphics[height=3.3cm]{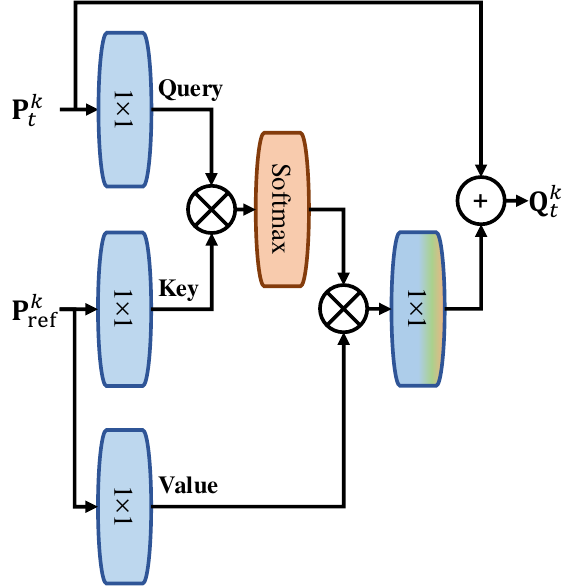}
    \label{fig:cross}
  }
  \hfill
  \subfloat[Spatiotemporal fusion module in PDNet.]{
    \centering
    \includegraphics[height=3.3cm]{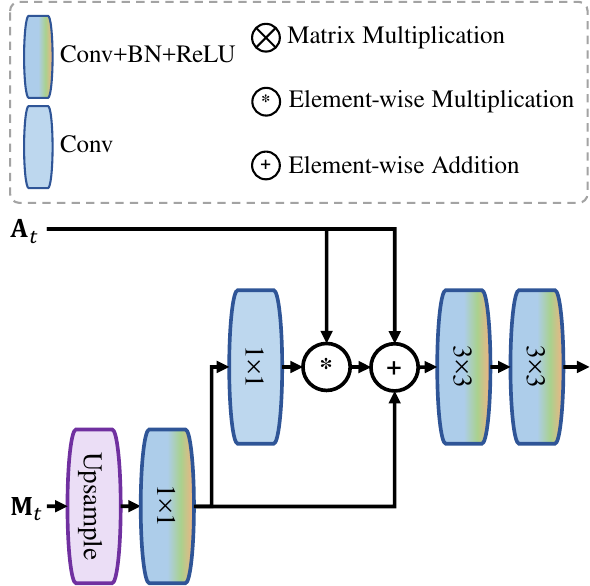}
    \label{fig:fusion}
  }
  \caption{Submodules in our proposed PDNet.}
  \label{fig:submodule}
\end{figure}

\paragraph{Unsupervised part decoupling}
In this stage, we unsupervisedly segment the portrait into several parts and extract the part-discriminated features simultaneously, which pipeline is shown in \cref{fig:PDStage}. Different from usual part decoupling methods~\cite{SCOPS, PartDAS}, in order to extract part-discriminated features that can contribute to portrait segmentation, we enhance the guidance of saliency cue in part segmentation. Specifically, we decompose the process of unsupervised part segmentation into two steps: First supervisedly predict the portrait saliency map based on appearance features, and then self-supervisedly predict each part of the portrait.

We first downsample appearance features $\mathbf{A}_t, \mathbf{A}_\text{ref}$ extracted by the Siamese encoder, and use a 3\texttimes 3 convolution operation with a 1\texttimes 1 convolution operation which compresses dimensions of $\mathbf{A}_t$ and $\mathbf{A}_\text{ref}$ to reduce the computational cost, and then a saliency predictor is applied to predict the portrait saliency map $\mathbf{S}_t$ and $\mathbf{S}_\text{ref}$. Then, in order to extract discriminated features from each portrait part, we utilize $p$ groups of a 1\texttimes 1, a 3\texttimes 3 and a 1\texttimes 1 convolution operations to decouple compressed features into part-discriminated features $\{\mathbf{P}_t^k\}_{k=0}^{p-1}$ and $\{\mathbf{P}_\text{ref}^k\}_{k=0}^{p-1}$, where $p$ denotes the number of parts. Stacked 1\texttimes 1 and 3\texttimes 3 convolution operations can enhance cross-channel information interaction and introduce more non-linearity, thus enhancing distinctiveness between part-discriminated features. Then, a weight-shared prediction head is applied on $p$ part-discriminated features to generate decoupled part predictions $\{\mathbf{D}_t^k\}_{k=0}^{p-1}$ and $\{\mathbf{D}_\text{ref}^k\}_{k=0}^{p-1}$ under the constraint of self-supervision losses: geometric concentration loss $\mathcal{L}_\text{geo}$, semantic consistency loss $\mathcal{L}_\text{sem}$ and area variance loss $\mathcal{L}_\text{area}$, which are introduced in \cref{sec:loss}. This weight-sharing mechanism ensures semantic consistency among part-discriminated features. Predicted part masks $\{\hat{\mathbf{R}}_t^k\}_{k=0}^{p-1}, \{\hat{\mathbf{R}}_\text{ref}^k\}_{k=0}^{p-1}$ are obtained by multiplying portrait saliency map $\mathbf{S}_t, \mathbf{S}_\text{ref}$ and decoupled part predictions $\{\mathbf{D}_t^k\}_{k=0}^{p-1}, \{\mathbf{D}_\text{ref}^k\}_{k=0}^{p-1}$. Take the target frame for example, $\{\hat{\mathbf{R}}_t^k\}_{k=0}^{p-1}$ can be formulated as
\begin{equation}
  \hat{\mathbf{R}}_t^k=\mathbf{S}_t\odot\mathbf{D}_t^k, k\in\{0, 1, \cdots, p-1\},
\end{equation}
where $\odot$ denotes element-wise production. Operations for the reference frame are the same. Note that all of the operations in part decoupling stage are weight-shared between the features of target and reference frame, which ensures the correspondence of parts and the consistency of features between the two frames.

\paragraph{Correlation extraction}
After obtaining part-discriminated features $\{\mathbf{P}_t^k\}_{k=0}^{p-1}$ and $\{\mathbf{P}_\text{ref}^k\}_{k=0}^{p-1}$, we apply $p$ cross-attention modules~\cite{Transformer} on the part-discriminated features of $p$ parts separately to obtain part-discriminated motion features $\{\mathbf{Q}_t^k\}_{k=0}^{p-1}$ of the target frame. Note that different from general non-local self-attention module~\cite{non-local}, the cross-attention module we utilized is asymmetric, in which the query features are from discriminated features $\mathbf{P}_t^k$ of some part $k$ in the target frame, while the key and value features are from discriminated features $\mathbf{P}_\text{ref}^k$ of the corresponding part $k$ in the reference frame, as shown in \cref{fig:cross}. Part-discriminated motion correlations from parts in the reference frame to the corresponding parts in the target frame are separately extracted in this way.

\paragraph{Correlation assembly}
As shown in \cref{fig:PDNet}, in this stage we assemble the $k$ part-discriminated motion features $\{\mathbf{Q}_t^k\}_{k=0}^{p-1}$ of the target frame to acquire integrated motion features $\mathbf{M}_t$ by masking them with corresponding part masks $\{\hat{\mathbf{R}}_t^k\}_{k=0}^{p-1}$ and adding up them, which is formulated as
\begin{equation}
  \mathbf{M}_t=\sum_{k=0}^{p-1}\mathbf{Q}_t^k\odot\hat{\mathbf{R}}_t^k.
\end{equation}

\subsection{Spatiotemporal fusion}
\label{sec:fusion}
As both appearance cue in spatial domain and motion cue in temporal domain are important for video segmentation, $\mathbf{A}_t$ and $\mathbf{M}_t$ are fused through our spatiotemporal fusion module. The pipeline is shown in \cref{fig:fusion}, where we first use a 1\texttimes 1 convolution operation to expand the dimension of $\mathbf{M}_t$. After that, we use another 1\texttimes 1 convolution operation on the expanded motion features and multiply it with $\mathbf{A}_t$ to obtain spatiotemporal features, in which regions that are significant in both appearance and motion are more responsive. At last, we sum up the three features and use two 3\texttimes 3 convolution operations to generate fused features.

\subsection{Loss functions}
\label{sec:loss}
\paragraph{Losses for portrait segmentation}
We utilize weighted binary cross-entropy (weighted BCE) loss and L1 loss for portrait segmentation, which are commonly used in VOS methods~\cite{AGNN, COSNet}. We utilize these two losses on both the final prediction and other auxiliary outputs from the Siamese encoder, which can enhance portrait feature extraction ability of the encoder.

\paragraph{Losses for part decoupling stage in IPDA module}
In the saliency prediction step, we utilize weighted BCE loss and L1 loss as well on $\mathbf{S}_t$ and $\mathbf{S}_\text{ref}$. In the self-supervised part prediction step, we utilize geometric concentration loss $\mathcal{L}_\text{geo}(\{\mathbf{X}^k\})$ and semantic consistency loss $\mathcal{L}_\text{sem}(\{\mathbf{X}^k\}, \mathbf{Y})$ as introduced in~\cite{SCOPS}, where $\mathbf{X}^k$ denotes a predicted mask of part $k$ and $\mathbf{Y}$ denotes the mask of portrait. Additionally, we introduce the area variance loss $\mathcal{L}_\text{area}(\{\mathbf{X}^k\})$ for self-supervision, which aims to prevent generating oversized or undersized parts which may result in reducing the ability of IPDA module to resolve motion, which is defined as
\begin{equation}
  \mathcal{L}_\text{area}(\{\mathbf{X}^k\})=D(\{\mathrm{area}(\mathbf{X}^k)\}/\max(\lVert\{\mathrm{area}(\mathbf{X}^k)\}\rVert_2, \varepsilon)),
\end{equation}
where $D(\cdot)$ indicates the variance, $\varepsilon$ denotes a small constant that prevents division by zero, and $\mathrm{area}(\mathbf{X}^k)$ denotes the expected pixel number of the predicted part mask $\mathbf{X}^k$, which is computed as
\begin{equation}
  \mathrm{area}(\mathbf{X}^k)=\sum_{i=0}^{H-1}\sum_{j=0}^{W-1}\mathbf{X}^k(i, j).
\end{equation}

The constraint of $\mathcal{L}_\text{geo}$ ensures that pixels within each part tend to converge towards a specific region, enabling the model to generate parts with well-organized shape connectivity. Under the supervision of $\mathcal{L}_\text{sem}$, pixels with similar semantics tend to be assigned to the same part, while the discrepancies of semantic features between different parts are enlarged. This enables the model to distinguish portrait parts with different features more reasonably. $\mathcal{L}_\text{area}$ suppresses the generation of oversized or undersized parts, preventing the wasteful computation of these ineffective parts as the number of parts is limited. With the help of these three losses, reasonable parts can be generated to enhance the model's ability to extract part-discriminated motion correlations.

As the predicted saliency maps $\mathbf{S}_t, \mathbf{S}_\text{ref}$ are not accurate and stable enough for self-supervision on part segmentation, we also introduce ground truth masks $\mathbf{G}_t, \mathbf{G}_\text{ref}$ in the training stage and produce part masks $\{\mathbf{R}_t^k\}_{k=0}^{p-1}, \{\mathbf{R}_\text{ref}^k\}_{k=0}^{p-1}$ masked by the ground truths, which can be formulated as
\begin{equation}
  \begin{aligned}
  \mathbf{R}_t^k&=\mathbf{G}_t\odot\mathbf{D}_t^k, k\in\{0, 1, \cdots, p-1\}
  \end{aligned}
\end{equation}
for the target frame and the same operation for the reference frame.

Meanwhile, as the ground truths are not available during inference time, to enhance consistency between the training and inference time while ensuring accuracy, we use both $\{\hat{\mathbf{R}}_t^k\}_{k=0}^{p-1}, \{\hat{\mathbf{R}}_\text{ref}^k\}_{k=0}^{p-1}$ and $\{\mathbf{R}_t^k\}_{k=0}^{p-1}, \{\mathbf{R}_\text{ref}^k\}_{k=0}^{p-1}$ for self-supervision. The summed-up part segmentation loss $\mathcal{L}_{\text{total\_part}}$ of the target frame can be formulated as
\begin{equation}
  \mathcal{L}_{\text{total\_part}}=(\mathcal{L}_{\text{part}}(\{\mathbf{R}_t^k\}, \mathbf{G}_t)+\mathcal{L}_{\text{part}}(\{\hat{\mathbf{R}}_t^k\}, \mathbf{S}_t))/2,
\end{equation}
where
\begin{equation}
  \begin{aligned}
  \mathcal{L}_{\text{part}}(\{\mathbf{X}_t^k\}, \mathbf{Y}_t)=&\mathcal{L}_\text{geo}(\{\mathbf{X}_t^k\})+\mathcal{L}_\text{sem}(\{\mathbf{X}_t^k\}, \mathbf{Y}_t)\\ 
  &+\mathcal{L}_\text{area}(\{\mathbf{X}_t^k\}).
  \end{aligned}
\end{equation}
$\mathcal{L}_{\text{total\_part}}$ for the reference frame is the same.

\begin{table}[t]
  \footnotesize
  \centering
  \caption{Quantitative comparison of our proposed PDNet with 7 state-of-the-art unsupervised VOS methods on MVPS dataset. The best results in each column are shown in \textbf{bold}.}
  \label{tab:comparison}
  \begin{tabular}{c|c|c|C{.7cm}|C{.7cm}|C{.8cm}|C{.8cm}|C{.7cm}|C{.8cm}|C{.8cm}|c}
    \toprule
    \multirow{2}{*}{Method} & \multirow{2}{*}{Origin} & \multirow{2}{*}{Backbone} & \multirow{2}{\linewidth}{\centering $\mathcal{J}$\&$\mathcal{F}$ Mean\textuparrow} & \multicolumn{3}{c|}{$\mathcal{J}$} & \multicolumn{3}{c|}{$\mathcal{F}$} & \multirow{2}{*}{FPS}\\
    \cline{5-10}
     & & & & Mean\textuparrow & Recall\textuparrow & Decay\textdownarrow & Mean\textuparrow & Recall\textuparrow & Decay\textdownarrow &\\
    \midrule
    COSNet~\cite{COSNet} & CVPR 2019 & ResNet-101 & 85.7 & 87.8 & 97.7 & 2.4 & 83.5 & 98.2 & 1.8 & 2.5\\
    AGNN~\cite{AGNN} & ICCV 2019 & ResNet-101 & 80.6 & 83.1 & 93.5 & 2.4 & 78.1 & 93.3 & 2.0 & 1.2\\
    MATNet~\cite{MATNet} & TIP 2020 & ResNet-101 & 79.2 & 83.1 & 94.6 & 1.2 & 75.2 & 92.0 & \textbf{0.9} & 10.2\\
    F2Net~\cite{F2Net} & AAAI 2021 & ResNet-101 & 77.5 & 82.2 & 93.1 & 2.1 & 72.9 & 86.4 & 2.9 & 11.1\\
    FSNet~\cite{FSNet} & ICCV 2021 & ResNet-50 & 84.2 & 86.5 & 95.8 & 2.0 & 81.9 & 97.5 & 1.8 & 23.6\\
    AMC-Net~\cite{AMC-Net} & ICCV 2021 & ResNet-101 & 85.9 & 88.1 & 98.1 & \textbf{0.6} & 83.6 & 97.4 & 1.1 & 22.8\\
    HFAN-medium~\cite{HFAN} & ECCV 2022 & MiT-B2 & 86.9 & 88.6 & 96.9 & 2.7 & 85.1 & 95.5 & 1.2 & 10.2\\
    IMCNet~\cite{IMCNet} & TCSVT 2022 & ResNet-101 & 84.1 & 86.6 & 97.2 & 2.0 & 81.6 & 97.6 & 1.3 & 23.9\\ 
    \midrule
    PDNet & Ours & ResNet-50 & \textbf{88.1} & \textbf{90.0} & \textbf{98.3} & 2.2 & \textbf{86.1} & \textbf{98.3} & \textbf{0.9} & \textbf{26.6}\\
    \bottomrule
  \end{tabular}
\end{table}

\begin{table}[t]
  \footnotesize
  \centering
  \caption{Quantitative comparison on PP-HumanSeg14K~\cite{PP-HumanSeg} dataset. The best results in each column are shown in \textbf{bold}.}
  \label{tab:comparison-pp}
  \begin{tabular}{c|c|c|c}
    \toprule
    Method & $\mathcal{J}$\&$\mathcal{F}$ Mean & $\mathcal{J}$ Mean & $\mathcal{F}$ Mean\\
    \midrule
    COSNet~\cite{COSNet} & 95.7 & 96.9 & 94.6\\
    AGNN~\cite{AGNN} & 93.9 & 96.1 & 91.6\\
    MATNet~\cite{MATNet}  & 93.1 & 95.6 & 90.7\\
    F2Net~\cite{F2Net} & 89.3 & 93.9 & 84.7\\
    IMCNet~\cite{IMCNet} & 94.5 & 96.4 & 92.7\\
    \midrule
    PDNet & \textbf{96.0} & \textbf{97.1} & \textbf{94.9}\\
    \bottomrule
  \end{tabular}
\end{table}

\begin{figure}[t]
  \centering
  \includegraphics[width=\linewidth]{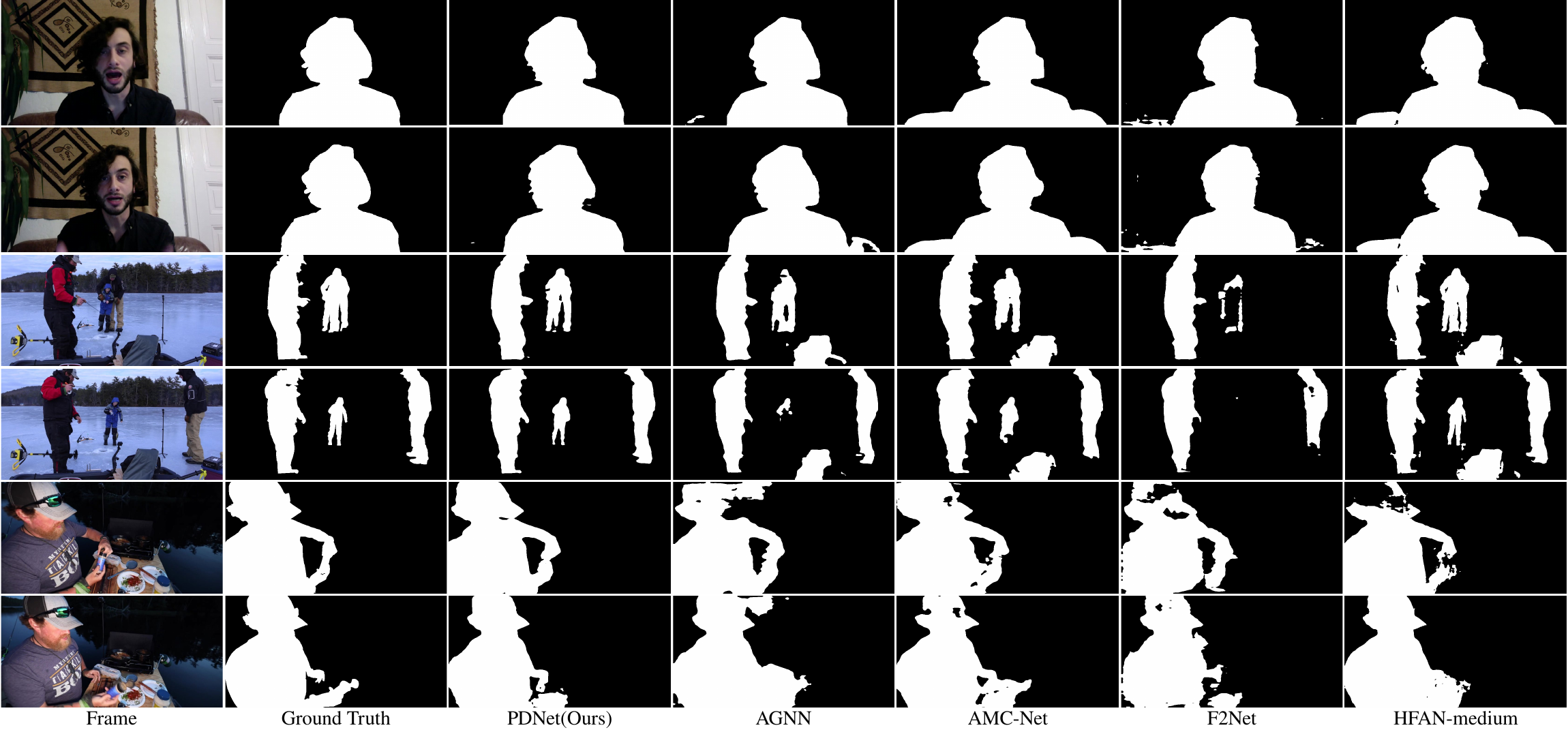}
  \caption{Visual comparison between our method and state-of-the-art methods on MVPS dataset.}
  \label{fig:qualitative}
\end{figure}

\section{Experiments}
\subsection{Implementation details}
We implement our network by PyTorch~\cite{PyTorch} and use two NVIDIA RTX 3090 GPUs for conducting our experiments. The number of parts $p$ in the IPDA module is experimentally set to $5$. ResNet-50~\cite{ResNet} pre-trained on ImageNet~\cite{ImageNet} is adopted as the backbone. As for semantic consistency loss $\mathcal{L}_\text{sem}$, we adopt a weight-fixed ResNet-18~\cite{ResNet} to generate semantic features for parts, which is not used during inference. Besides MVPS dataset, we also use several image portrait segmentation datasets described in \cref{sec:dataeval} to fine-tune the backbone as the same training strategy in~\cite{AGNN, COSNet, F2Net}. That is, images and video frames are used alternately during training, where $1$ iter of backbone fine-tuning by images is conducted before per $2$ iters of training by video frames. All of the images, target frames and reference frames fed into the network are randomly cropped, randomly flipped in horizontal direction and resized to a fixed size of $480\times 480$ during training. Stochastic gradient descent (SGD) optimizer with a momentum of $0.9$ and a weight decay of $5\times 10^{-4}$ is used for training. The size of mini-batch is set to $4$ and the number of epochs is set to $60$. During fine-tuning by image, the maximum learning rate is set to $2.5\times 10^{-4}$, while during training by video frames it is set to $2.5\times 10^{-5}$ for backbone and $2.5\times 10^{-3}$ for the rest of our model. During inference, both target and reference frames are directly resized to $480\times 480$ for prediction.

\subsection{Datasets and evaluation metrics}
\label{sec:dataeval}
We use our proposed MVPS dataset for training and evaluating our network and state-of-the-art unsupervised VOS methods. As for the image datasets used for fine-tuning the backbone, we use images in EG1800~\cite{EG1800} (1,735 images available), SuperviselyPerson (5,711 images available) and MSCOCO~\cite{MSCOCO}. For the MSCOCO dataset, we select images including persons of which the pixel number is no less than 5,000, and pick out 49,479 images. For fair comparison, the same image set is also used during training on VPS for those methods that utilize image datasets during training on VOS. For evaluation metrics, we adopt region similarity $\mathcal{J}$ and contour accuracy $\mathcal{F}$ which are commonly used in VOS task~\cite{DAVIS}.

\subsection{Comparison results}

\paragraph{Quantitative comparison on MVPS}
We compare our proposed PDNet with 8 state-of-the-art unsupervised VOS methods including COSNet~\cite{COSNet}, AGNN~\cite{AGNN}, MATNet~\cite{MATNet}, F2Net~\cite{F2Net}, FSNet~\cite{FSNet}, AMC-Net~\cite{AMC-Net}, HFAN-medium~\cite{HFAN} and IMCNet~\cite{IMCNet} which codes are publicly available for retraining and evaluation. For fair comparison, a unified threshold of 0.5 is utilized on predictions from all methods to generate binary portrait masks, while CRF is not used for post-processing.

The quantitative results of our PDNet and other 7 methods mentioned above are shown in \cref{tab:comparison}. From the results we can see that our PDNet achieves significantly leading performance with the comparison to state-of-the-art methods on the test set of MVPS, which outperforms the second best method HFAN-medium by $1.4\%$ and $1.0\%$ on $\mathcal{J}$ Mean and $\mathcal{F}$ Mean, respectively. Note that our PDNet utilizes ResNet-50 as backbone, while most state-of-the-art unsupervised VOS methods utilize ResNet-101~\cite{ResNet} with more parameters and stronger feature extraction capability.

We also compare the efficiency of our PDNet with state-of-the-art methods under the same environment, as shown in the last column of \cref{tab:comparison}. Our method achieves a speed of 26.6 FPS, which is the most competitive among these methods.

\begin{figure}[t]
  \centering
  \includegraphics[width=0.53\linewidth]{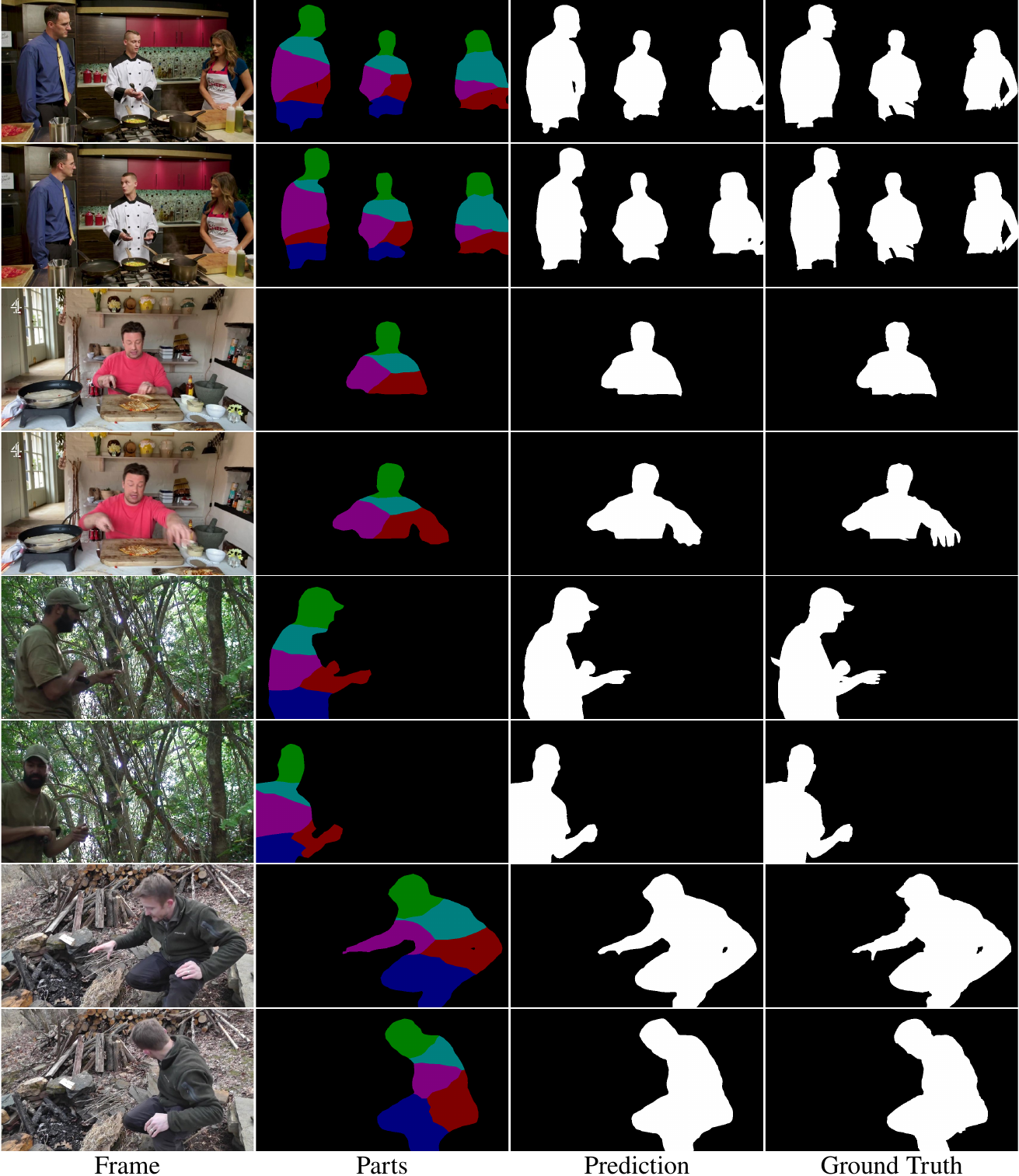}
  \caption{Visual examples of decoupled parts and segmentation results of our PDNet on the test set of MVPS dataset.}
  \label{fig:part}
\end{figure}

\paragraph{Quantitative comparison on PP-HumanSeg14K}
To further demonstrate the robustness of our method and the complexity of our MVPS dataset, we also train and evaluate our PDNet and several state-of-the-art unsupervised VOS methods on PP-HumanSeg14K~\cite{PP-HumanSeg} dataset. The comparison results are shown in \cref{tab:comparison-pp}. Our approach also achieves the best performance. Note that the overall performance on PP-HumanSeg14K dataset is significantly higher than that on MVPS dataset, which shows our proposed MVPS dataset is more challenging due to the complexity of our dataset.

\paragraph{Qualitative comparison}
To intuitively demonstrate the excellent performance of our method, we compare the visualization results of our method with state-of-the-art VOS methods on the test set of our MVPS dataset. As shown in \cref{fig:qualitative}, our method segments foreground portraits accurately in a variety of portrait numbers, portrait motion, portrait gestures and background scenes, especially in regions with imbalanced motion against the main bodies, while other methods incorrectly predict objects in backgrounds such as sofa, parcel, \etc. as foreground portrait area, or miss some regions of portraits even the whole portrait of someone. Although our method has achieved good performance, there is still room for improvement. For instance, the segmentation results of boundary areas or regions near occluded objects are not yet satisfactory. We will focus on further enhancing the model's ability to handle details in our future work.

\subsection{Visual analysis of decoupled parts}
As an important cue, part plays a crucial role in our method. Here we show some visual examples of decoupled parts as intermediate results in \cref{fig:part}. Under the effect of self-supervision, our method decouples portraits into parts which are usually with different motion states. Portrait parts generated by our IPDA module are robust to the variation of background scenes and portrait motion, even when there are multiple persons in one video clip. Based on this robustness, our method can extract discriminated motion correlation of different portrait parts separately to capture imbalanced motion of portraits. Note that although the current segmentation result is not perfect, it can already provide important auxiliary information for imbalanced motion extraction, which improves the accuracy of final segmentation results.

\subsection{Ablation study}

\begin{table}[t]
  \footnotesize
  \centering
  \caption{Ablation study on the IPDA module and self-supervision strategies in part decoupling on MVPS dataset.}
  \label{tab:part-ablation}
  \begin{tabular}{C{0.65cm}|C{0.81cm}C{0.81cm}C{0.81cm}|C{0.8cm}|C{0.7cm}|C{0.7cm}}
    \toprule
    \multirow{2}{*}{Base} & \multicolumn{3}{c|}{Self-supervision Losses} & \multirow{2}{\linewidth}{\centering $\mathcal{J}$\&$\mathcal{F}$ Mean} & \multirow{2}{\linewidth}{\centering $\mathcal{J}$ Mean} & \multirow{2}{\linewidth}{\centering $\mathcal{F}$ Mean}\\
    \cline{2-4}
    & $\mathcal{L}_\text{geo}$ & $\mathcal{L}_\text{area}$ & $\mathcal{L}_\text{sem}$ & & & \\
    \midrule
    \checkmark & \multicolumn{3}{c|}{w/o IPDA and fusion} & 85.0 &  88.2 & 81.9\\
    \checkmark & \checkmark & & & 86.5 & 89.0 & 84.0\\
    \checkmark & \checkmark & \checkmark & & 87.3 & 89.4 & 85.2\\
    \midrule
    \checkmark & \checkmark & \checkmark & \checkmark & \textbf{88.1} & \textbf{90.0} & \textbf{86.1}\\
    \bottomrule
  \end{tabular}
\end{table}

\begin{figure}[t]
  \centering
  \includegraphics[width=0.56\linewidth]{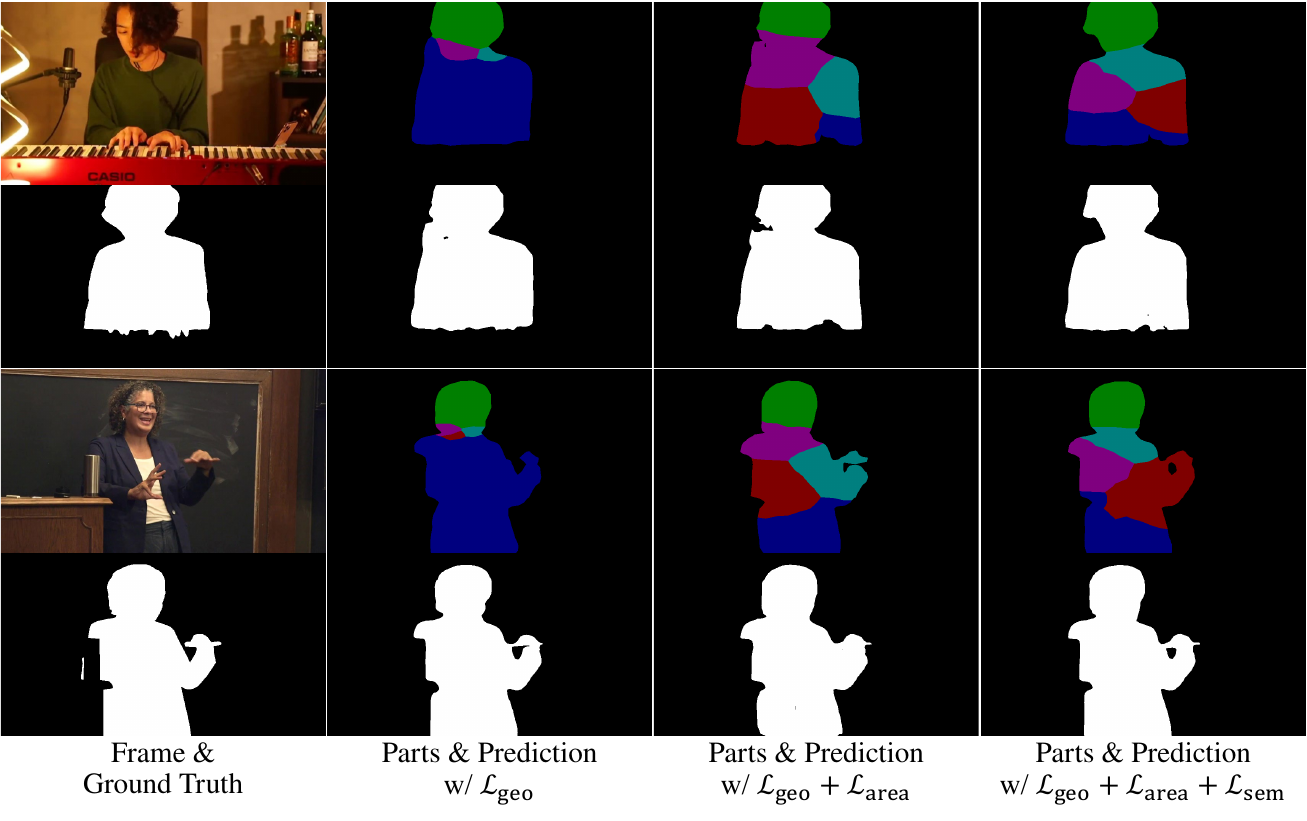}
  \caption{Visual comparison for ablation study on self-supervision losses in part decoupling. Frames and decoupled parts are shown in odd lines, while ground truths and segmentation results are shown in even lines.}
  \label{fig:part-ablation}
\end{figure}

\paragraph{Impact of the IPDA module}
To further verify the effectiveness of our IPDA module and self-supervision strategies in part decoupling for video portrait segmentation, we conduct experiments as shown in \cref{tab:part-ablation}. In the first row, we remove the whole IPDA and spatiotemporal fusion modules, while in the second row we replace part segmentation losses with only $\mathcal{L}_\text{geo}$. It can be seen that introducing rough part cue can also improve the performance even without area and semantic supervision. In the third row, we reserve $\mathcal{L}_\text{geo},\mathcal{L}_\text{area}$ and remove $\mathcal{L}_\text{sem}$, which demonstrates the effectiveness of our proposed area variance loss. It prevents oversized or undersized parts from being generated, which guarantees stability of the ability of IPDA module to decompose portrait motion. From the third and the last row, we can see that semantic cue is also important to enhance the motion-resolving ability, as owing to the joint structure of human body, parts with different appearance semantics are often in different motion states. From \cref{fig:part-ablation} we can see that under the guidance of more appropriate part maps generated by adding $\mathcal{L}_\text{area}$ and $\mathcal{L}_\text{sem}$ during training, the model can segment part boundary regions and regions with obvious motion more accurately.

\begin{table}[t]
  \footnotesize
  \centering
  \caption{Ablation study on the number of parts $p$ in our IPDA module on MVPS dataset.}
  \label{tab:partnum}
  \begin{tabular}{C{1.5cm}|c|c|c}
    \toprule
    Number of Parts $p$ & $\mathcal{J}$\&$\mathcal{F}$ Mean & $\mathcal{J}$ Mean & $\mathcal{F}$ Mean\\
    \midrule
    1 & 86.2 & 88.8 & 83.6\\
    3 & 86.7 & 88.7 & 84.7\\
    \textbf{5} & \textbf{88.1} & \textbf{90.0} & \textbf{86.1}\\
    \bottomrule
  \end{tabular}
\end{table}

\begin{table}[!t]
   \footnotesize
   \centering{
   \caption{Ablation study on the number of reference frames of our PDNet on MVPS dataset.}
   \label{tab:refnum}
   \begin{tabular}{C{2.3cm}|c|c|c|c}
    \toprule
    Number of Reference Frames & $\mathcal{J}$\&$\mathcal{F}$ Mean & $\mathcal{J}$ Mean & $\mathcal{F}$ Mean & FPS\\
    \midrule
    1 & 88.1 & 90.0 & 86.1 & 26.6\\
    2 & 88.7 & 90.7 & 86.8 & 9.7\\
    \bottomrule
   \end{tabular}
  }
\end{table}

\paragraph{Impact of the number of parts $p$}
To verify the impact of the number of parts $p$ in our IPDA module, we conduct experiments as shown in \cref{tab:partnum}. Note that when the number of parts is set to 1, losses for self-supervised part segmentation are removed as they are not needed. From the table, we can see that within a certain range, the segmentation performance improves as the number of parts increases. This is because the increase in the number of parts enables portrait regions with different motion states to be separated while extracting motion correlation, which enhances the decoupling ability of the model to complex motion. As we use NVIDIA RTX 3090 GPU for experiments, the memory is almost full during training when the number of parts is set to 5, so we use 5 as the final setting of number of parts in our method.

\paragraph{Impact of the number of reference frames}
To verify the impact of the number of reference frames, we add the fifth annotated frame before the current frame as the second reference frame, and the experimental results are shown in \cref{tab:refnum}. From the table we can see that although there is an improvement in performance, there is a significant decline in the efficiency of the model, with FPS dropping from 26.6 to 9.7, a decrease of 63.5\%. The involvement of more convolution and attention operations significantly increases computational overhead, resulting in a decrease in speed. Since the balance between performance and efficiency is important, and using 1 reference frame is a typical setting in video segmentation methods~\cite{AnDiff,F2Net,PVS}, we use 1 reference frame in our method.

\section{Conclusion}
In this paper, we propose a new intricate large-scale dataset MVPS and a Part-Decoupling Network for video portrait segmentation. The proposed dataset MVPS addresses the lack of intricate large-scale multi-scene VPS datasets, which contains 101 video clips and 10,843 annotated frames in 7 categories of scenarios. It is currently the most complex dataset for VPS. Towards the imbalance of portrait motion, our proposed Part-Decoupling Network decouples portrait motion into part-level. In our network, the Inter-frame Part-Discriminated Attention module is proposed to extract discriminative part-level motion features from the target and reference frames. Experimental results demonstrate that our method achieves superior performance with the comparison to state-of-the-art unsupervised VOS methods.

\end{document}